# RealMedQA: A pilot biomedical question answering dataset containing realistic clinical questions

Gregory Kell, MPhil[1], Angus Roberts, PhD[1], Serge Umansky, PhD[2], Yuti Khare, MBBS[3], Najma Ahmed, BSc, MBBS[1], Nikhil Patel[1], Chloe Simela, MD[1], Jack Coumbe[1], Julian Rozario[1], Ryan-Rhys Griffiths, PhD[4], Iain J. Marshall, PhD, MD[1],

[1]King's College London, London, Greater London, United Kingdom; [2]Metadvice Ltd., London, Greater London, United Kingdom; [3]Maidstone and Tunbridge Wells NHS Trust, Maidstone, Kent, United Kingdom; [4]University of Cambridge, Cambridgeshire, United Kingdom

**Abstract**

*Clinical question answering systems have the potential to provide clinicians with relevant and timely answers to their questions. Nonetheless, despite the advances that have been made, adoption of these systems in clinical settings has been slow. One issue is a lack of question-answering datasets which reflect the real-world needs of health professionals. In this work, we present RealMedQA, a dataset of realistic clinical questions generated by humans and an LLM. We describe the process for generating and verifying the QA pairs and assess several QA models on BioASQ and RealMedQA to assess the relative difficulty of matching answers to questions. We show that the LLM is more cost-efficient for generating "ideal" QA pairs. Additionally, we achieve a lower lexical similarity between questions and answers than BioASQ which provides an additional challenge to the top two QA models, as per the results. We release our code[1] and our dataset[2] publicly to encourage further research.*

## Introduction

Clinical question answering (QA) systems could allow clinicians to find timely and relevant answers to questions occurring during consultations in real-time [1, 2, 3, 4, 5]. This is especially important given the vast amount of information available to clinicians, and the time constraints they face [6, 7, 8].

Despite the progress that has been made, biomedical question answering systems are still not widely used in practice. Even though they have seen continual improvement on benchmarks, the answers generated by these systems are not useful to clinicians [9]. The reasons for this include the fact that the answers do not originate from reliable sources of information, the answers are not in the form of guidance, they do not account for the clinician's setting, the rationale provided for the answers is not sufficient, and that systems do not resolve and communicate conflicting evidence and uncertainties adequately.

The capabilities of the systems are, in part, limited by the characteristics of existing biomedical QA datasets and benchmarks. Common practices include deriving questions from the titles of research articles [10] and employing a diverse biomedical expert panel [11]. Questions are often designed to test the systems' ability to comprehend biomedical corpora, instead of eliciting information that would be required by clinicians in practice [12, 13]. Additionally, the quality of the source texts in corpora is often not verified, meaning that they could be out of date or contain biased results. While these biomedical QA datasets may be suitable for investigating novel methods, they may not be appropriate for systems intended for clinical settings. Thus, there is a need for a dataset that consists of realistic clinical questions, as well as answers that have been verified by medical professionals.

To address this research gap, we present RealMedQA: a realistic biomedical QA dataset where the answers consist of clinical guideline recommendations provided by the UK-based National Institute for Health and Care Excellence (NICE). The questions were generated by a group of medical students and by a large language model (LLM). The quality of all the question-answer pairs then underwent verification conducted by the same group of medical students. We show the LLM is more cost-efficient for generating "ideal" QA pairs. Several QA models were tested on the RealMedQA dataset and BioASQ [11] to compare the relative difficulty of matching the questions to the correct answers. This quantitative investigation is supplemented by a qualitative comparison of a sample of BioASQ [11] and RealMedQA QA pairs.

---

[1] https://github.com/gck25/RealMedQA
[2] https://huggingface.co/datasets/k2141255/RealMedQA

Our key contributions are summarized as follows:

- We present RealMedQA: a biomedical dataset consisting of clinical questions that are realistic and whose answers are clinical guideline recommendations, thus satisfying the requirements for reliability, guidance and rationale.
- We provide a comparison of the questions generated by humans vs those generated by a LLM and show that the LLM is more cost-efficient.
- We provide the results of both quantitative and qualitative comparisons which suggest that the QA pairs of RealMedQA have a lower lexical similarity than those of BioASQ.

**Methods**

We followed a SQuAD-style [14, 15] process to generate the question answer pairs, where we used guideline recommendations from the UK National Institute for Health and Care Excellence (NICE) as the answers. The process can be broken down into three stages: **data collection**, **question generation** and **question-answer pair verification**. Our methodology resembles that of SQuAD; however, our objective diverges by focusing on the creation of question-answer pairs for information retrieval rather than extractive or span-based QA. Considering the increasing prevalence of generative AI in biomedical QA [13], the relevance of extractive QA may diminish over time. Consequently, our dataset prioritizes information retrieval. Furthermore, this dataset is specifically designed for the primary care/general medicine domain, where clinical guideline recommendations are generally sufficient.

The **data collection** stage involved downloading the guidelines via the NICE syndication[3] API and extracting the recommendations from the guidelines. Next, we hired six UK-based medical students via a university employment agency to **generate questions** that were addressed by the recommendations. The experience of the students varied from first year of (undergraduate) medical school to first-year medical junior doctors in the UK's National Health Service. Up to 5 questions were created by the human generators for each recommendation and up to 20 were generated by the LLM (gpt-3.5-turbo[4]). Finally, the quality of the dataset was assured through a round of **verification** by four of the medical students.

*Data collection*

All the guidelines available via the NICE syndication API were downloaded for further processing. The total number of guidelines was 12,543, grouped in several high-level topics. As we were focusing on guidelines for care provision (as opposed to administration), we limited the high-level topics to "Conditions and diseases" which reduced the number of guidelines to 7,385.

*Question generation*

We created an instruction sheet on formulating questions for each recommendation. After extraction, this was used by the medical students to develop questions. Questions were then collected using Google Forms[5]. The generators were asked to provide up to 5 questions for each recommendation where the emphasis was on questions that clinicians would be likely to ask during clinical practice. The aim was to avoid comprehension-style questions such as "Who should be included in a core specialist heart failure multidisciplinary team?" and "What are the responsibilities of the specialist heart failure MDT?", as it is unlikely that clinicians would be interested in these types of questions. Rather, research shows that they would be more likely to ask questions in the style of "What is the drug of choice for condition X?" or "What is the cause of symptom X?"[14], i.e. questions that pertain to the diagnosis, treatment and management of patients.

To ensure generators' understanding of the task, we provided several examples of questions created for given recommendation. For example, when given with the recommendation "Offer an angiotensin-converting enzyme (ACE) inhibitor and a beta-blocker licensed for heart failure to people who have heart failure with reduced ejection

---

fraction. Use clinical judgement when deciding which drug to start first. [2010]", potential questions answered by the recommendation could be:

- What treatment should I use for heart failure?
- Which groups with heart failure should have a beta blocker?
- Which drug should I use first in drug failure?

These training examples were verified by one of the authors of the paper (Iain J. Marshall) who is a primary care physician. To ensure that the generators understood the instructions and the rationale behind these examples, several meetings were organized to discuss the instructions and answer any questions that occurred.

The generators were directed to the Google Drive directory with all the forms organized under categories, e.g. 'Cancer', and sub-categories, e.g., 'Ovarian cancer'. They were then asked to submit questions for recommendations for a wide range of categories. We permitted questions from multiple question generators per recommendation for comparison, but we asked the generators to limit overlaps to 10 forms. The question generators were asked to provide up to 5 questions per recommendation. A total of 1150 QA pairs were generated using this method.

In addition to the human question generators, an LLM was used to generate physician's questions. GPT3.5 Turbo was accessed using the API provided by OpenAI[6]. The default temperature parameter of 1.0 was used to balance the diversity and coherence of the answers. To generate questions for a given recommendation two prompts were submitted to the LLM: one consisting of the recommendation itself and the other consisting of the text in the instruction sheet used by the question generators. The LLM was prompted to provide 20 questions per recommendation instead of 5. In total, 100325 question-recommendation pairs were generated.

*Question-answer pair verification*
During the QA pair verification phase, we created Google forms with all the QA pairs generated by the human question generators and some generated by the LLM. The verifiers were then asked to answer the following questions using a Likert scale for each QA pair:

- Does the question look like the sort of question a doctor would ask in practice? (**Plausible**)
- Is the question answered by the guideline? (**Answered**)

Each of the questions could be answered with the following options: "Completely", "Partially" and "Not at all". The verifiers were four of the same students who generated questions in the previous phase; they would verify the questions generated by the LLM and each other. We ensured that no one was asked to verify the questions that they themselves created. To address potential discrepancies among verifiers, we conducted an initial pilot study to identify the sources and reasons for such discrepancies. Based on the pilot results, we revised the instructions to minimize future discrepancies. We henceforth refer to question that were rated as both "completely" plausible and "completely" answered by the recommendations as "ideal".

Once the instruction sheet was updated, each verifier was allocated 200 QA pairs created by humans and 100 generated by the LLM. As there were four verifiers, a total of 800 human QA pairs and 400 LLM QA pairs were verified (1200 in total). Blinding was applied to the source of each QA pair.

In addition to the verification matrix, we also report the inter-verifier agreement scores between two verifiers based on 50 human QA pairs and 50 LLM QA pairs. We calculated the agreement for both the plausibility of the questions and whether they are answered by the recommendations. The scores include the following:

- **Cohen's kappa**, where "Completely", "Partially" and "Not at all" ratings are treated as separate categories (Table 2);
- **Cohen's kappa**, where "Partially" and "Not at all" ratings are treated as one category (Table 3);

---

[6] https://platform.openai.com/

We use the aforementioned scores to compare the relative agreement over whether or not a question was "completely" or is and the more fine-grained categories. This would demonstrate whether the verifiers were more likely to disagree on a binary or fine-grained classification and whether a more consistent means of defining each rating would be necessary going forward.

*Experiments*

To assess the difficulty of matching RealMedQA's questions and answers relative to other datasets, we evaluated several question-and-answer encoders on all the "ideal" QA pairs. Cosine similarity was used to assess the similarity between question-and-answer vectors. To assess the quality of the QA matchings, we used recall@k where $k$ was set to 1, 2, 5, 10, 20, 50 and 100, as well as nDCG@k and MAP@k where $k$ was set to 2, 5, 10, 20, 50 and 100. The recall@k is the fraction of correct answers included in the top $k$ list, while the nCDG@k evaluates the ordering of the top $k$ list (more relevant items should have higher confidence scores). The MAP@k also assesses the quality of the ranked top $k$ list. We plot the 95% Wald confidence interval [17] associated with recall@k and MAP@k, as those could be interpreted as the probability of correctly identifying a positive given a set of true positives and predicted positives, respectively. The nDCG, on the other hand, does not represent a probability, so it would not be meaningful to calculate the Wald confidence interval for this metric. We compared several models, including BM25 (term frequency-based ranking function; [18, 19]) and BERT-based LLMs (110 million parameters): the original BERT (pre-trained on Wikipedia data; [20]), SciBERT (fine-tuned on scientific articles; [21]), BioBERT (fine-tuned on biomedical articles; [22]), PubMedBERT (pre-trained on biomedical articlecls; [23]) and Contriever (fine-tuned using a contrastive loss; [24]). Although BERT has demonstrated suboptimal performance on semantic encoding tasks without additional training or pooler representations [25], there is currently no equivalent of Contriever or sentence-based encodings specifically for the biomedical domain (hence the development of biomedical LLMs). Furthermore, our dataset was insufficiently large to fine-tune any of the models. Therefore, we included the general BERT model as a baseline for comparison with the biomedical variants. We also evaluated the same models on a BioASQ-based retrieval dataset [11] used as part of BEIR [26] to compare its difficulty with that of RealMedQA; we used the titles and abstracts for the BioASQ answers. We randomly sampled the same number of QA pairs from BioASQ as there were in the "ideal" RealMedQA dataset. We compliment the quantitative comparison of RealMedQA and BioASQ with a qualitative investigation of the QA pairs. We used the Gensim and Rank-BM25 Python packages to conduct the BM25 experiments. The BERT-based evaluations utilized the transformers, datasets, torch and faiss-gpu packages. We employed the scikit-learn implementation of all the metrics. The experiments were carried out on a 24GB NVIDIA GEFORCE RTX 3090 and each run took fewer than 10 minutes.

**Results**

*Plausibility and answerability*

Out of 1200 QA pairs, only 230 (19% (2 s.f.)) were deemed to contain questions that were "completely" plausible with recommendations that "completely" addressed the questions (Table 1).

**Table 1.** Matrix showing number of QA pairs corresponding to **plausibility** and **answered** rating ("Completely", "Partially", "Not at all") during QA pair verification phase.

| Answered/ Plausible | Completely | Partially | Not at all |
|---|---|---|---|
| **Completely** | 230 | 181 | 228 |
| **Partially** | 44 | 90 | 122 |
| **Not at all** | 21 | 41 | 243 |

*Inter-verifier agreement*

The inter-verifier agreement scores are shown in Tables 2 and 3. The inter-verifier agreement scores are consistently higher for the human QA pairs than for those of the LLM (where the effects are especially pronounced for the combined Cohen's kappa in Table 3). Nonetheless, the inter-verifier agreement is substantially higher for the **plausibility** scores than the **answered** scores. The Cohen's kappa in Table 3 is lower for the LLM's "answered" ratings (0.22) than Table 2 (0.24), suggesting that the reviewers disagreed more over whether the LLM "completely" answered the questions than whether the LLM's answer fitted into the fine-grained category. This demonstrates the lack of clarity over what constitutes a "completely" answered question, especially when it was generated by an LLM.

**Table 2.** Cohen's kappa for the verification, where the categories were "Completely", "Partially", "Not at all".

| Question generators | Plausible | Answered |
|---|---|---|
| Humans | 0.64 | 0.30 |
| LLM | 0.60 | 0.24 |

**Table 3.** Cohen's kappa for the verification, where the categories "Partially" and "Not at all" were grouped together.

| Question generators | Plausible | Answered |
|---|---|---|
| Humans | 0.85 | 0.37 |
| LLM | 0.59 | 0.22 |

*Costs for generating and verifying questions*

Table 4 shows the costs of the human question generators, the number of QA pairs generated and the cost per QA pair in dollars. The cost of the humans was taken to be the wages paid, while the LLM cost was the cost of calling the OpenAI API. Table 5 also shows the cost for verifying each QA pair based on estimates obtained during the pilot verification phase. The main cost for the LLM is the question verification, while that of the humans is the question generation. Additionally, before accounting for question quality, it is approximately three times more expensive to generate question answer pairs using humans than the LLM. We account for question quality in our cost calculations in Table 5. Humans are shown to be more efficient than LLMs with a "ideal" QA yield of 0.222, compared with 0.130 for the LLM. Moreover, when we account for question quality, the cost per "ideal" QA pair of the LLM is just over half that of the humans.

**Table 4.** The total cost, the number of QA pairs and the cost per QA pair for the LLM and human generators. Note: the human generators were paid in GBP, while access to the LLM was paid for in USD. We converted the prices to USD using Google finance on 19th October 2023 at 13:30.

| Question generators | Total cost ($; 3 s.f.) | Number of QA pairs (3 s.f.) | Question generation cost ($) per QA pair (3 s.f.) | Question verification cost ($) per QA pair (3 s.f.) | Total cost ($) per QA pair (3 s.f.) |
|---|---|---|---|---|---|
| Humans | 2930 | 1150 | 2.55 | 1.21 | 3.76 |
| LLM | 14.0 | 100000 | 0.0000140 | 1.21 | 1.21 |

**Table 5.** Comparison QA pairs generated by LLM and by human question generators. "Ideal" QA pairs are those that are both "completely" plausible and whose recommendations "completely" addresses the question.

| Question generators | Total number of QA pairs | Number of "ideal" QA pairs | "Ideal" QA pair yield (3 s.f.) | Cost ($) per QA pair (3 s.f.) | Cost ($) per "ideal" QA pair (0 d.p.) |
|---|---|---|---|---|---|
| Humans | 800 | 178 | 0.222 | 3.76 | 17 |
| LLM | 400 | 52 | 0.130 | 1.21 | 9 |

*Quantitative comparison of BioASQ and RealMedQA*

As shown in **Figure 1A** the recall@k of BM25 on BioASQ is greater than that of Contriever for lower values of $k$, while Contriever matches the performance of BM25 for higher values. On the other hand, Contriever consistently outperforms BM25 on RealMedQA (**Figure 1B**). The confidence intervals do not overlap for higher values of k on RealMedQA which means that the differences are statistically significant. Furthermore, both BM25 and Contriever perform worse on RealMedQA than on BioASQ, while the reverse is true for the other models. All the BERT-based models apart from Contriever perform worse than BM25. This suggests that the self-supervised losses used to train these models are not sufficient for discerning the relative similarity of text spans, even when the models are pre-trained/fine-tuned on domain-specific data. The recall@k is consistently greater for BERT than all the fine-tuned scientific and biomedical models on BioASQ. In contrast, the recall@k is consistently higher for BioBERT than BERT on RealMedQA, while the recall@k of SciBERT matches that of BERT. BioBERT fine-tuned on PubMedQA [10] outperforms BERT on RealMedQA for lower values of $k$, while the opposite is true for higher values of $k$. BioBERT fine-tuned on PubMedQA under-performs when compared to the original on both BioASQ and RealMedQA. PubMedBERT has a consistently higher recall@k than the BioBERT variant fine-tuned on PubMedQA on BioASQ, while PubMedBERT has the lowest recall@k on RealMedQA for all values of $k$.

**Figure 1.** Recall@k for BioASQ (**A**) and b) recall@k for RealMedQA (**B**) for *k* equal to 1, 2, 5, 10, 20, 50 and 100. The shaded areas represent the 95% Wald confidence interval.

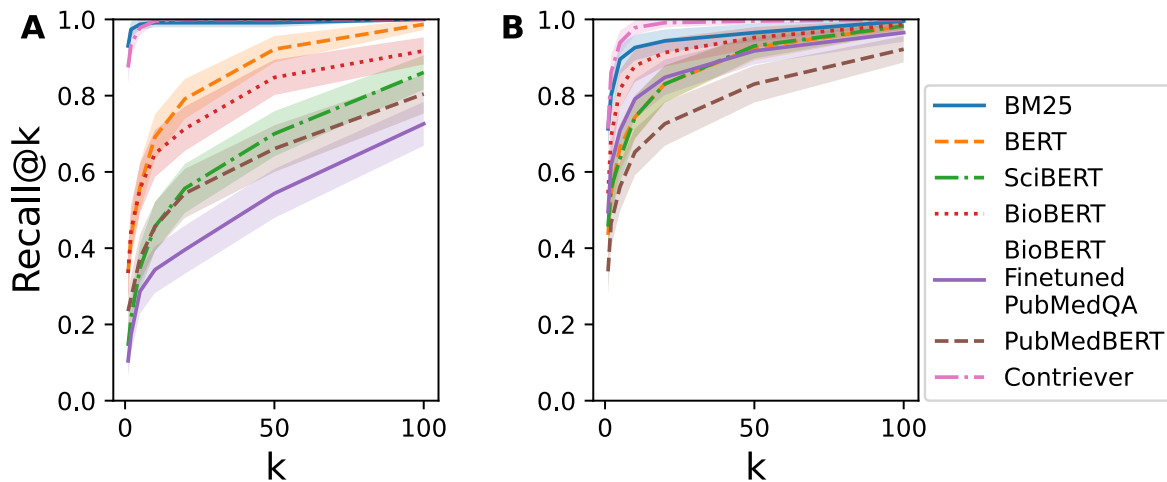

MAP@k is consistently greater for Contriever than for BM25 on both BioASQ (**Figure 2A**) and RealMedQA (**Figure 2B**). Nonetheless, both Contriever and BM25 perform worse on RealMedQA than on BioASQ, while the converse is true for the other models. Additionally, the performance of BM25 worsens as *k* increases, in contrast to the other models. Similarly, to recall@k, BERT has a higher MAP@k on BioASQ than the other biomedical and scientific variants, while it is outperformed on RealMedQA by both BioBERT variants. SciBERT matches the MAP@k of BERT for all values of *k*. The original BioBERT consistently outperforms the BioBERT variant fine-tuned on PubMedQA on both BioASQ and RealMedQA. PubMedBERT consistently outperforms BioBERT fine-tuned on PubMedQA and SciBERT on BioASQ, while it is outperformed by all models but BM25 on RealMedQA. The same results are seen for nDCG@k (**Figure 3**).

**Figure 2.** MAP@k for BioASQ (**A**) and MAP@k for RealMedQA (**B**) for k equal to 2, 5, 10, 20, 50 and 100. The shaded areas represent the 95% Wald confidence interval.

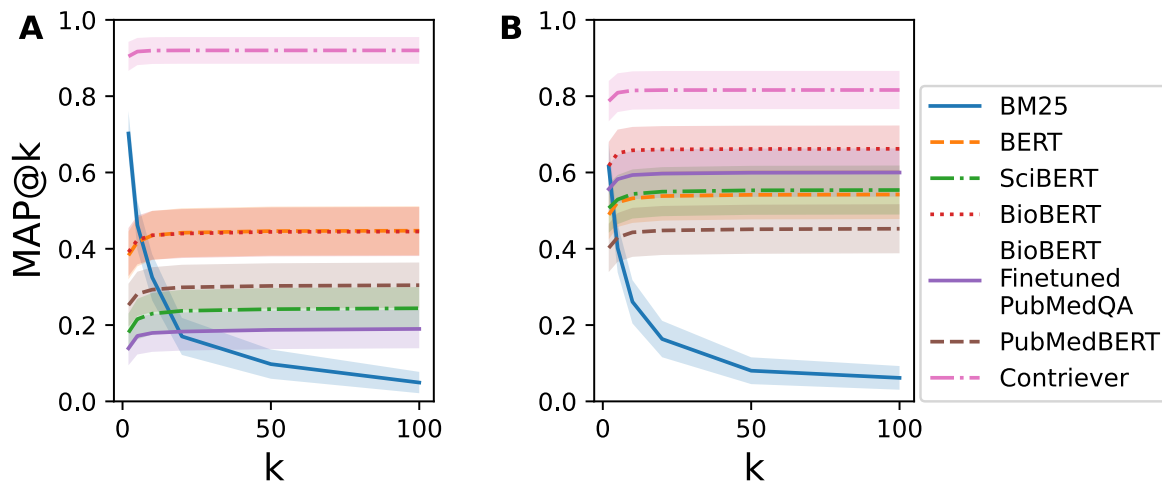

**Figure 3.** nDCG@k for BioASQ (**A**) and nDCG@k for RealMedQA (**B**) for k equal to 2, 5, 10, 20, 50 and 100. The Wald confidence interval was not calculated for nDCG@k, as nDCG@k cannot be interpreted as a probability.

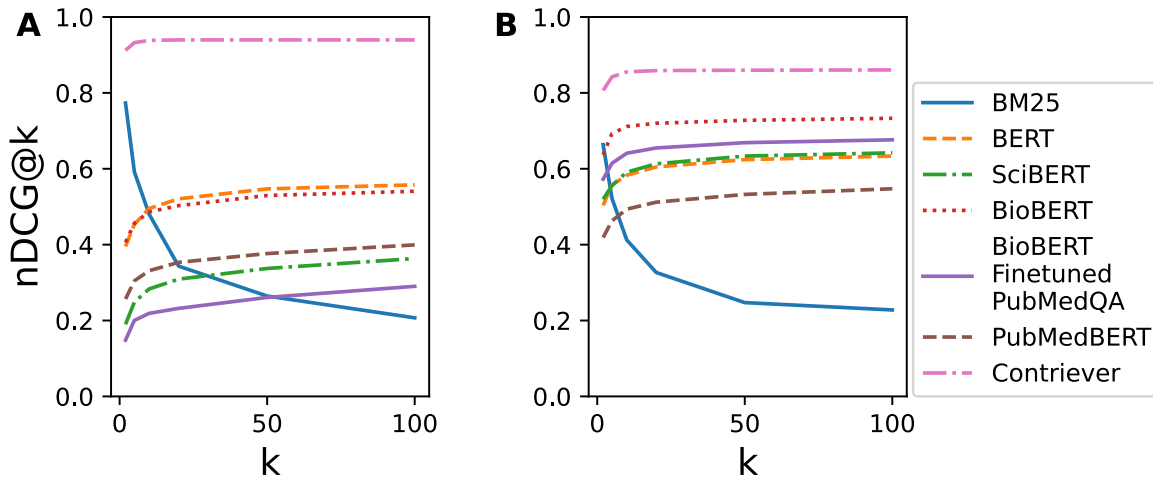

*Qualitative comparison of RealMedQA and BioASQ*

To account for the difference of the models on each dataset we examined QA pairs randomly sampled from each dataset. Within the BioASQ dataset, the question "dna methylation in tongue is not found in the blood" is accompanied by the title and abstract of the study with PMID 28603561 [27]. The title is "Distinct DNA methylation profiles in subtypes of orofacial cleft." Upon inspection, there appears to be a high lexical similarity between the question and the title and abstract. For example, the word "methylation" is in the title and recurs regularly in the abstract, while the same can be said for "blood". "DNA" also appears both in the title and the abstract.

On the other hand, the question "What considerations should be addressed while revising the management programme for a child or adolescent?" from RealMedQA is accompanied by the following recommendation from the guideline "Spasticity in under 19s: management": "If adverse effects (such as drowsiness) occur with oral diazepam or oral baclofen, think about reducing the dose or stopping treatment."

When comparing the two samples, there is a greater lexical overlap between the question and answer from BioASQ than between those of RealMedQA. Furthermore, BioASQ was not intended for clinical use which is why the dataset consists of general basic science questions, while RealMedQA contains specific clinical questions that elicit information of interest to the clinician. Given that the intention of BioASQ is to assess the performance of QA systems in the biomedical domain, fewer checks would have been required for the reliability of the answer compared with the NICE guidelines.

**Discussion**

*LLM vs human QA pair quality*

In the previous sections, we compared the yield for "ideal" QA pairs of the LLM and human question generators, where we showed that the human yield was 7.4% greater than that of the LLM. In other words, LLMs are competitive with humans, especially when the cost per QA pair is considered. Nonetheless, the low inter-verifier agreement over whether the questions were adequately answered by the recommendations (even with examples in the instructions), demonstrates the subjective nature of QA pair generation. Further research is required on the standardization of QA pairs.

*LLM vs human QA pair costs*

While the employment of LLMs does bring down the cost of question generation significantly, $12 is still required for each "ideal" QA pair, due to verification costs. This means that even generating a QA dataset with 1000 QA pairs

would cost $12,000. This cost could be reduced using automatic verification methods, although, as previously stated, further work would be required to standardize the definition of "ideal" QA pairs.

*Comparison of RealMedQA and BioASQ datasets*

The superior performance of BM25 to the other models on BioASQ suggests a high degree of lexical similarity between the questions and answers. On the other hand, the improved performance of Contriever compared with other models on RealMedQA implies a lower degree of lexical similarity between the questions and answers. This finding is supported by the random samples taken from BioASQ and RealMedQA.

Additionally, BERT's superior performance to all the other models on BioASQ shows that in-domain fine-tuning is useful only for RealMedQA and not for BioASQ. The importance of general domain data for pretraining is highlighted by PubMedBERT's low performance on both datasets, as PubMedBERT was only pretrained on PubMed abstracts. This is especially important when dealing with other forms of biomedical data such as clinical guidelines.

Fine-tuning BioBERT on PubMedQA does not improve the performance on either task. As PubMedQA questions are derived from the titles of the source papers, the high lexical similarity between the questions and the answers could hinder the performance of the fine-tuned model on both datasets.

*Comparison with other biomedical QA datasets*

Several biomedical QA datasets have been developed for various purposes, including benchmarking the knowledge encoded by LLMs [28], reading comprehension over electronic health records [29], and reading comprehension over biomedical article abstracts [10]. While these datasets assess the quality of LLMs and biomedical QA components (information extractors), our dataset aims to retrieve clinically approved guideline recommendations using realistic clinical questions, addressing an existing research gap [12, 13].

*Limitations*

Since the experiments were conducted, newer versions of GPT (e.g., 4 and 4o; [30]) have been released. It is important to note that generative AI is a rapidly evolving field, and the cost analysis might change in light of these updates. Furthermore, we recognize that retrieval-augmented generation (RAG) workflows [31, 32] may provide more appropriate answers to clinicians. However, this work focuses solely on the retrieval of clinical guideline recommendations using realistic clinical questions.

*Potential implications and future directions*

The dataset introduced in this paper consists of realistic clinical questions created by medical students and a large language model (LLM). We have demonstrated the potential of LLMs to create machine learning datasets, considering quality and verification costs. We envision that our findings will lead to the extended use of LLMs for dataset creation, including in the biomedical question-answering (QA) domain. Although the dataset is currently too small for training purposes, we aim to expand it by leveraging LLMs to create additional questions, with medical professionals verifying the QA pairs. We also plan to explore the use of novel prompting techniques, including self-verification of the questions and QA pairs via language model cascades [33]. The extended dataset would provide a more realistic representation of biomedical QA for clinicians and could drive methodological breakthroughs in this area, leading to systems that better address clinicians' information needs. Additionally, the dataset could be extended to address both generation and retrieval for RAG.

## Conclusion

In this work, we have presented a methodology for generating a biomedical question answering dataset containing realistic clinical questions and answers. We compared LLM and human question generators and have found that LLMs are more cost-efficient, even when QA quality is considered. Nonetheless, the LLM's cost-efficiency is heavily affected by the human scrutiny applied to the output of the LLM. Furthermore, we found that human verifiers are unlikely to agree on whether a question has been adequately answered by a recommendation. Finally, the experimental results suggest that the lexical similarity between the questions and answers of RealMedQA are lower than that of BioASQ. Hence, the methodology presented here offers a promising avenue to creating biomedical QA datasets.